\documentclass{article}
\usepackage[margin=1in]{geometry} 
\usepackage{PRIMEarxiv}
\usepackage{amsmath}
\usepackage[utf8]{inputenc} 
\usepackage[T1]{fontenc}    
\usepackage{hyperref}       
\usepackage{url}            
\usepackage{booktabs}       
\usepackage{amsfonts}       
\usepackage{microtype}      
\usepackage{fancyhdr}       
\usepackage{graphicx}       
\usepackage{xcolor}
\usepackage{float}
\usepackage{tabularx} 
\usepackage{subcaption}
\graphicspath{{figures/}}

\usepackage{xspace}

\newcommand{\best}[1]{\textcolor{green!50!black}{#1}}
\newcommand{\worst}[1]{\textcolor{red}{#1}}

\begin{document}

\title{FCMBench-Video: Benchmarking Document Video Intelligence} 


\author{
  \textbf{
    Runze Cui\textsuperscript{1},
    Fangxin Shang\textsuperscript{1},
    Yehui Yang\textsuperscript{1 \thanks{Corresponding author. Contact: yangyehuisw@126.com}},
    Qing Yang\textsuperscript{1},
    Yanwu Xu\textsuperscript{3,4},
    Tao Chen\textsuperscript{2}, }\\
  \textsuperscript{1} AI Lab, Qifu Technology, Beijing, China \\
  \textsuperscript{2} College of Future Information Technology, Fudan University, Shanghai, China \\
  \textsuperscript{3} School of Future Technology, South China University of Technology, Guangzhou, China \\
  \textsuperscript{4} Pazhou Lab, Guangzhou, China \\
}

\maketitle

\begin{abstract}
Document understanding is a critical capability in financial credit review, onboarding, and remote verification, where both decision accuracy and evidence traceability matter. 
Compared with static document images, document videos present a temporally redundant and sequentially unfolding evidence stream, require evidence integration across frames, and preserve acquisition-process cues that are useful for authenticity-sensitive and anti-fraud review. 
We introduce \textbf{FCMBench-Video}, a benchmark for document-video intelligence that evaluates document perception, temporal grounding, and evidence-grounded reasoning under realistic capture conditions. 
To support privacy-compliant yet realistic data at scale, we organize benchmark construction as an atomic-acquisition and composition workflow that records reusable single-document clips, applies controlled degradations, and assembles long-form multi-document videos with prescribed temporal spans. 
FCMBench-Video is a bilingual benchmark built from \textbf{495} captured atomic videos, composed into \textbf{1,200} long-form videos and paired with \textbf{11,322} expert-annotated question--answer instances. It covers \textbf{28} document types over duration tiers from \textbf{20s} to \textbf{60s}, including \textbf{5,960} Chinese instances and \textbf{5,362} English instances.
Evaluations on nine recent Video-MLLMs show that FCMBench-Video provides meaningful separation across systems and capabilities: counting is the most duration-sensitive task, Cross-Document Validation and Evidence-Grounded Selection probe higher-level evidence integration, and Visual Prompt Injection provides a complementary robustness dimension. The overall score distribution is broad and approximately bell-shaped, indicating that the benchmark is neither saturated nor dominated by trivial cases. Together, these results position FCMBench-Video as a reproducible benchmark for tracking Video-MLLM progress on document-video understanding and for probing capability boundaries in authenticity-sensitive credit-domain applications.

We release FCMBench-Video and the evaluation protocol at \href{https://github.com/QFIN-tech/FCMBench}{this URL}.

\end{abstract}

\begin{figure*}[!htbp]
    \centering
    \includegraphics[width=0.8\linewidth]{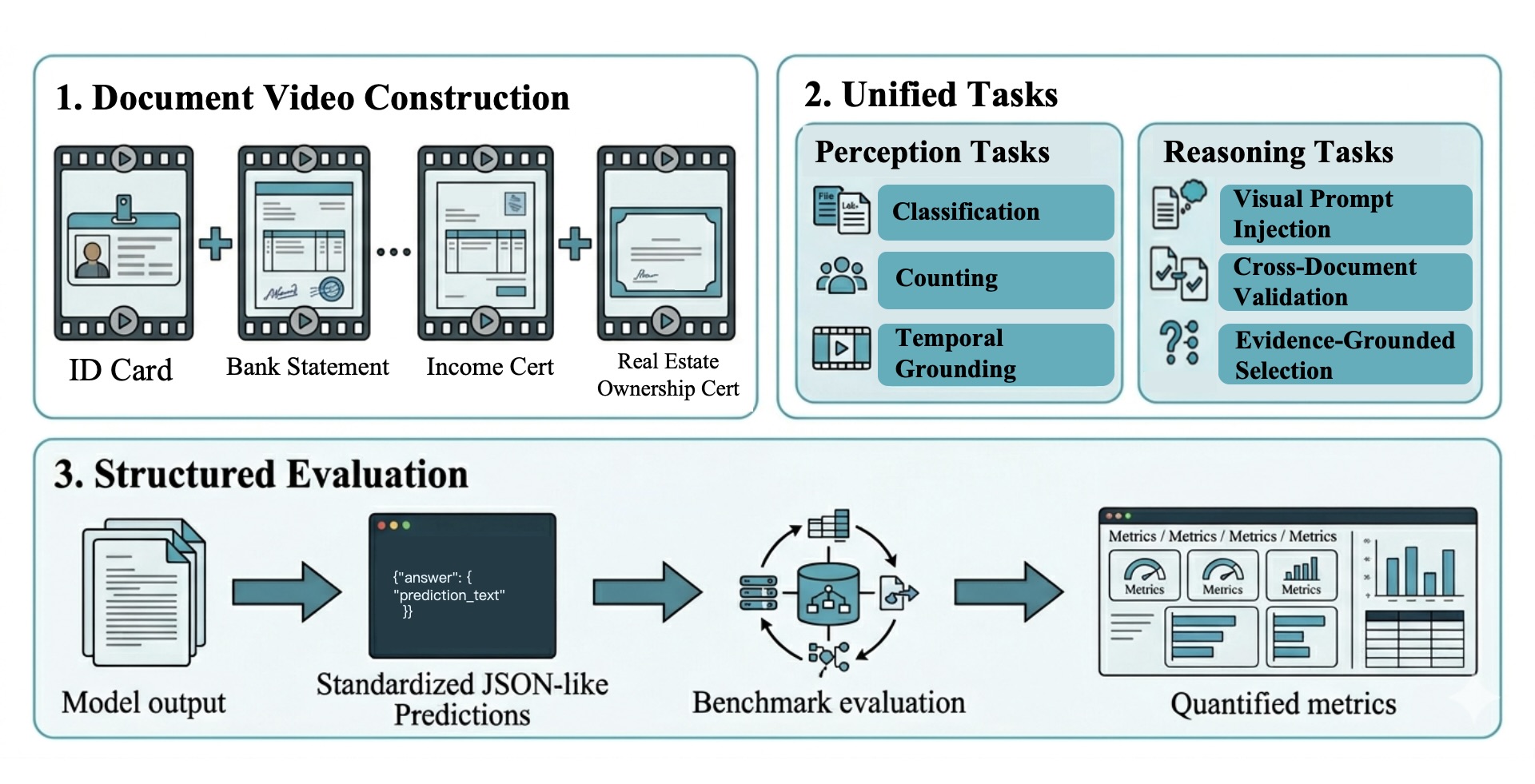}
    \caption{Overview of FCMBench-Video. A document video is represented as a temporally ordered stack of oblique frame slices to emphasize continuity rather than isolated snapshots. From this shared video input, the benchmark derives a unified set of perception and reasoning tasks, including Classification, Counting, Temporal Grounding, Visual Prompt Injection, Cross-Document Validation, and Evidence-Grounded Selection. Model outputs are then converted into structured predictions for reproducible evaluation.}
    \label{fig:teaser}
\end{figure*}

\section{Introduction}   
\label{sec:introduction}

Document understanding is a critical capability in financial credit review and related real-world workflows, where systems must interpret document evidence, support auditable decisions, and preserve evidence traceability. 
Previous work, FCMBench~\cite{yang2026fcmbenchcomprehensivefinancialcredit}, made important progress in this direction by establishing an image-based multimodal benchmark for financial credit document analysis under realistic workflow constraints. 
It provides a strong foundation for evaluating document perception, page-level reasoning, and robustness in static document settings.

However, many operational document-analysis scenarios are not limited to isolated images. 
In handheld scanning, onboarding, and remote verification, users often present, flip, and move documents in front of a camera, producing \emph{document videos} whose evidence unfolds over time. 
Unlike static document images, document videos present a temporally redundant and sequentially unfolding evidence stream, in which task-relevant information is not always concentrated in a single frame and may need to be integrated across time. This setting requires models to suppress uninformative frames, integrate complementary evidence over time, and preserve traceability through \emph{temporal grounding}. Moreover, the continuity of the acquisition process preserved in document videos provides richer signals for authenticity assessment and anti-fraud review than a small set of manually selected still images.
Video-capable multimodal large language models (Video-MLLMs) are increasingly used as practical visual interfaces, but evaluating them only on static document images leaves these video-specific capabilities under-specified.

Despite recent progress in video understanding benchmarks, existing evaluations remain insufficient for this setting. General-purpose video benchmarks are typically curated from web sources (e.g., YouTube) and emphasize coarse-grained semantics such as actions, events, and narratives, offering limited coverage of document recognition, cross-frame evidence aggregation, and audit-grade evidence alignment. Meanwhile, document understanding benchmarks are predominantly image-centric, focusing on single-page snapshots that cannot evaluate temporal continuity, document transitions, or cross-segment consistency. As a result, current leaderboards provide limited guidance for assessing whether Video-MLLMs can perform reliably on document-centric videos encountered in practice.

Building a benchmark for this setting is challenging. Real document videos often contain personally identifiable information (PII) and are rarely shareable, creating a fundamental tension between privacy compliance and scenario realism. Synthetic approaches that render documents or generate templated frames can improve compliance, but they often fail to reproduce the acquisition dynamics of handheld recording, such as entry and exit motions, brief windows of peak legibility, and device-dependent imaging artifacts. This gap makes it difficult to construct a benchmark that is simultaneously privacy-compliant, realistic, scalable, and amenable to controlled difficulty adjustment.

To address these challenges, we introduce \textbf{FCMBench-Video}, a benchmark for \textbf{document-video intelligence} designed to evaluate Video-MLLMs on document perception and evidence-grounded reasoning over real-world document videos. FCMBench-Video extends document evaluation from static snapshots to temporally unfolding evidence streams and complements static and multi-image document benchmarks by covering capabilities that arise specifically in the video modality. In particular, it targets temporal evidence localization, long-context document transitions, robustness to visually injected malicious instructions, and authenticity-sensitive cues preserved by the acquisition process itself. Figure~\ref{fig:teaser} provides an overview of the benchmark input, task families, and structured evaluation flow.

For privacy-compliant data construction at scale, we organize benchmark assembly as an atomic-acquisition and composition workflow over \emph{captured} recordings. The workflow proceeds in three stages: short single-document clips are recorded under realistic handheld capture, optional photometric, optical, and codec degradations are applied, and the resulting clips are concatenated into multi-document videos with prescribed temporal spans. We treat this workflow as a description of how the benchmark is built, not as a methodological contribution; its purpose is to keep data construction privacy-compliant and reproducible while preserving the interaction dynamics of real document recording.

Following the physical credit and informational document types in FCMBench~\cite{yang2026fcmbenchcomprehensivefinancialcredit}, FCMBench-Video covers \textbf{28} document types across bilingual \textbf{Chinese} and \textbf{English} settings. Table~\ref{tab:video_collection} summarizes the captured atomic video collection. From \textbf{495} privacy-compliant atomic recordings, we assemble \textbf{1{,}200} long-form multi-document videos organized into \textbf{20s/40s/60s} duration tiers, paired with \textbf{11{,}322} expert-annotated question--answer instances (\textbf{5{,}960} in Chinese and \textbf{5{,}362} in English). Degraded atomic clips are stochastically mixed into each composition to reflect realistic acquisition noise.

We evaluate nine recent Video-MLLMs, including several state-of-the-art systems, and observe clear capability separation across task families, duration tiers, and robustness settings. These results show that FCMBench-Video exhibits non-trivial variation in document-video perception, temporal evidence use, and reasoning under realistic capture conditions, and therefore supports analysis beyond leaderboard comparison.

\begin{table}[!htbp]
\centering
\caption{Compact summary of the captured atomic video collection. All videos are recorded indoors with natural lighting, clean backgrounds, and no occlusion or reflection, using smartphones as capture devices.}
\label{tab:video_collection}
\setlength{\tabcolsep}{4pt}
\footnotesize
\begin{tabularx}{\linewidth}{p{0.11\linewidth} p{0.10\linewidth} p{0.46\linewidth} p{0.11\linewidth} p{0.14\linewidth}}
\toprule
\textbf{Region} & \textbf{\#Types} & \textbf{Representative categories} & \textbf{\#Videos} & \textbf{Duration range} \\
\midrule
zh-CN & 20 & Real estate certificate, loan application form, household register, bank statement, business license & 251 & 1.57s--96.7s \\
en-US & 8 & Social Security document, mortgage form, property title report, bank statement, driver's license & 244 & 1.37s--128.37s \\
\bottomrule
\end{tabularx}
\end{table}

In summary, we present \textbf{FCMBench-Video}, a comprehensive benchmark for document-video intelligence that evaluates Video-MLLMs on both perception and reasoning tasks under realistic acquisition conditions, with explicit relevance to authenticity-sensitive and anti-fraud review. FCMBench-Video and the evaluation toolkit are publicly released to facilitate reproducible research and accelerate progress in document-centric video understanding.

\section{Related Benchmarks}
\label{sec:related_benchmarks}

\subsection{General Video Benchmarks for Video-MLLMs}
Recent benchmarks have substantially advanced the evaluation of video-capable multimodal models (Video-MLLMs) by covering diverse temporal skills and video modalities. MVBench~\cite{li2024mvbench} constructs a comprehensive collection of temporal understanding tasks by converting static tasks into dynamic ones and leveraging ground-truth annotations from multiple public video datasets. Video-MME~\cite{fu2024videomme} provides a systematic evaluation of video analysis with diverse video types, durations, and manually annotated questions, emphasizing comprehensive capability measurement across short to long videos. Video-MMMU~\cite{hu2025videommmu} focuses on knowledge acquisition from professional educational videos, evaluating perception, comprehension, and adaptation across multiple disciplines. 
Despite their value, these benchmarks are primarily curated from web or general-domain videos and focus on coarse-grained semantics (events, actions, narratives, or educational content). They do not explicitly target the unique challenges of \emph{document videos}, where document evidence is distributed across a redundant temporal stream, information must be integrated across partially informative frames, and \emph{temporal evidence grounding} is essential for auditability and authenticity-sensitive review.

\subsection{Document Visual Understanding Benchmarks}
Document understanding has been extensively studied in the image domain through benchmarks centered on reading, layout understanding, and key information extraction. DocVQA~\cite{mathew2021docvqa} evaluates question answering over document images, highlighting the need for structured document understanding beyond generic VQA. FUNSD~\cite{jaume2019funsd} targets form understanding with entity labeling/linking under noisy scanned conditions. The ICDAR SROIE competition report~\cite{huang2021sroie} provides datasets and protocols for receipt OCR and key information extraction. More recently, domain-specific benchmarks such as FCMBench~\cite{yang2026fcmbenchcomprehensivefinancialcredit} emphasize workflow relevance, privacy compliance, and robustness in financial credit document analysis under static single-image and multi-image settings.

These image-based settings remain important and are not superseded by FCMBench-Video. Instead, they anchor evaluation in document perception, page-level reasoning, and cross-image aggregation, while FCMBench-Video extends the same general line of inquiry to a complementary modality in which evidence is exposed over time through handheld capture. What static and multi-image benchmarks do not explicitly preserve is the acquisition process itself: entry and exit motions, fluctuating legibility, temporal evidence localization, and recency conflict between earlier and later visual content. Those properties become central once the input is a continuous document video rather than an unordered set of images, especially when the application cares not only about reading the document correctly but also about judging whether the capture process appears authentic.

\subsection{What Is Missing and How FCMBench-Video Fills the Gap}
Across existing benchmarks, there is a clear gap in standardized evaluation for \emph{document-video intelligence}. General video benchmarks rarely provide rigorous evaluation of document recognition, instance counting, and temporal evidence localization over redundant handheld capture streams, while document benchmarks are predominantly static and do not measure temporal grounding or long-range cross-segment consistency. FCMBench-Video targets this missing setting by benchmarking Video-MLLMs on document perception under realistic acquisition artifacts, temporal evidence grounding for traceability, and evidence-grounded reasoning over long video contexts. 
Moreover, unlike web-curated video benchmarks, FCMBench-Video is constructed from \textbf{captured} workflow-style document recordings under a privacy-compliant process, with controlled temporal scaling and stochastic mixing of degraded atomic clips. As detailed in Sec.~\ref{sec:adc}, each benchmark instance is organized along three axes---temporal span, degradation type, and composition structure---and videos are stratified over 20s--60s durations with atomic clips drawn from both Readable and Unreadable pools. This organization lets FCMBench-Video cover a wider range of document-video conditions within a single reproducible release than prior static document benchmarks or web-video leaderboards.

\section{Benchmark Construction}
\label{sec:adc}

FCMBench-Video targets \emph{document video} as a distinct input modality characterized by handheld acquisition dynamics and temporally unfolding document evidence. Constructing such a benchmark faces a practical tension between \emph{scenario realism} and \emph{privacy compliance}: real-world credit/workflow recordings often contain PII and are not shareable, while purely synthetic rendering tends to miss key acquisition artifacts and interaction dynamics. Realism matters not only for perception difficulty but also for application value, since in many remote review settings the capture process itself contains authenticity cues relevant to anti-fraud assessment. We address this tension with a three-stage \emph{Atomic--Degradation--Composition} (ADC) workflow: short single-document clips are recorded under realistic handheld capture, controlled degradations are optionally applied, and the resulting clips are composed into long-form multi-document videos with prescribed temporal spans.

\subsection{Workflow Overview}
\label{subsec:adc_overview}

\begin{figure}[!htbp]
    \centering
    \includegraphics[width=0.8\linewidth]{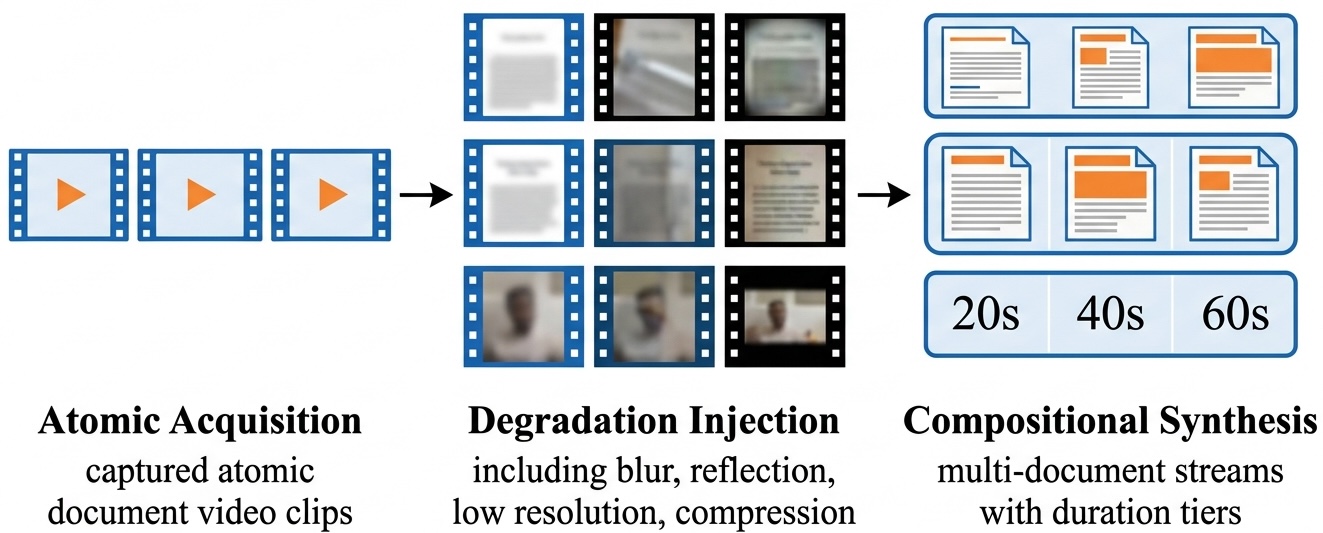}
    \caption{Atomic--Degradation--Composition (ADC) workflow for constructing privacy-compliant document videos. The workflow first records reusable atomic document--camera interactions under realistic handheld capture, then applies controlled degradations to simulate photometric, optical, and codec corruption, and finally composes the resulting clips into long-form multi-document videos with deterministic temporal annotations. Acquisition dynamics from the atomic stage are preserved, while corruption level, temporal duration, and evidence structure are controlled during composition.}
    \label{fig:adc_pipeline}
\end{figure}

The ADC workflow has three stages: (1) the \textbf{A}tomic Acquisition Stage collects short, single-document interaction clips as reusable units; (2) the \textbf{D}egradation Injection Stage applies controlled photometric/optical/codec perturbations to simulate real acquisition/transmission artifacts; and (3) the Video \textbf{C}omposition Stage assembles atomic units into coherent multi-document videos with deterministic temporal annotations. Fig.~\ref{fig:adc_pipeline} illustrates the overall workflow.

\subsection{Atomic Acquisition Stage}
\label{subsec:adc_atomic}

The Atomic Acquisition Stage is designed to preserve the physical capture characteristics of document videos while keeping the resulting clips reusable for later composition. To reflect real deployment conditions such as mobile remote verification, we record atomic clips with heterogeneous smartphones and preserve device-specific imaging pipeline characteristics, including sensor noise, ISP behavior, and color science. Although the target capture specification spans 720P--4K at 30/60\,FPS, we do not force visual normalization at acquisition time, since these device-dependent properties often affect practical document legibility. To ensure that subsequent evaluation remains focused on the target document, acquisition follows a strict unobstructed protocol: the recording area is kept free of underlying papers and extraneous clutter, thereby reducing inter-document textual crosstalk and spurious background text.

Each atomic clip also preserves the temporal structure of realistic handheld capture. Specifically, clips include natural \emph{in-and-out} phases in which the operator brings the document into the camera's field of view and removes it after recording, together with a short ``golden window''---typically a few seconds---during which the document is most legible. We retain both the golden window and the surrounding transition frames so that later compositions preserve realistic motion and legibility fluctuation rather than consisting only of manually selected clear frames. Acquisition is organized at the identity level: for each identity instance, collectors record all relevant certificates as a series of atomic clips, each corresponding to one document category. For multi-page documents, pages are captured sequentially within a single continuous atomic file so that intra-document temporal logic is preserved. To avoid geometric inconsistencies during later composition, a consistent orientation is maintained within each identity, while orientation is allowed to vary across identities as part of natural capture variation.

The output of this stage is a library of isolated atomic clips with structured metadata. Each clip is associated with a document type (one of 28 predefined categories), an identity ID for identity-level grouping and cross-document composition, device model and resolution, orientation (portrait/landscape), a readability label (Readable/Unreadable, assigned via 3-annotator consensus), and golden-window timestamps indicating the start and end of the most legible segment. These metadata support downstream composition, temporal supervision, and controllable benchmark instantiation.

\subsection{Degradation Injection Stage and Readability Labels}
\label{subsec:adc_degradation}

The Degradation Injection Stage introduces systematic and controllable corruption to emulate acquisition and transmission conditions that arise in real document-video workflows. The goal is not only to make document perception and temporal localization more realistic, but also to create benchmark instances with controlled variation in readability. We consider three main degradation families. First, for photometric interference, we simulate \emph{dynamic specular reflections} with a radial Gaussian model whose center evolves smoothly over time, and \emph{gradient shadow occlusion} with a smoothstep interpolation
$f(x)=3x^2-2x^3$ so that the resulting boundaries remain physically plausible rather than introducing hard synthetic edges. Second, for optical and geometric degradation, we apply isotropic Gaussian blur with varying $\sigma$ to mimic focus hunting or lens smearing, and downsample clips with bicubic interpolation to 480P as a representative low-resolution regime. Third, for codec degradation, we manipulate H.264/H.265 settings, including constant-bitrate throttling to 150 kbps (CBR) and high-compression settings with CRF=40, both of which introduce blocking, ringing, and loss of high-frequency detail that is important for document interpretation.

These degradations also support atomic-level readability labeling. At atomic granularity, we partition clips into \emph{Readable} and \emph{Unreadable} samples according to whether key document evidence remains recoverable to humans. Raw clips and moderate degradations such as photometric interference and 480P resampling are treated as Readable, whereas extreme blur or 150 kbps CBR are treated as Unreadable. The labels are verified by three independent annotators, and a sample is marked Unreadable only when all annotators agree that the primary document content cannot be reliably interpreted. These readability labels support atomic-level abstention and hallucination analysis on unreadable evidence, and also serve as a controlled source of quality variation for the later composition stage.

\subsection{Video Composition Stage: Temporal Annotations and Duration Tiers}
\label{subsec:adc_composition}

\begin{figure}[!htbp]
  \centering
  \includegraphics[width=0.7\linewidth]{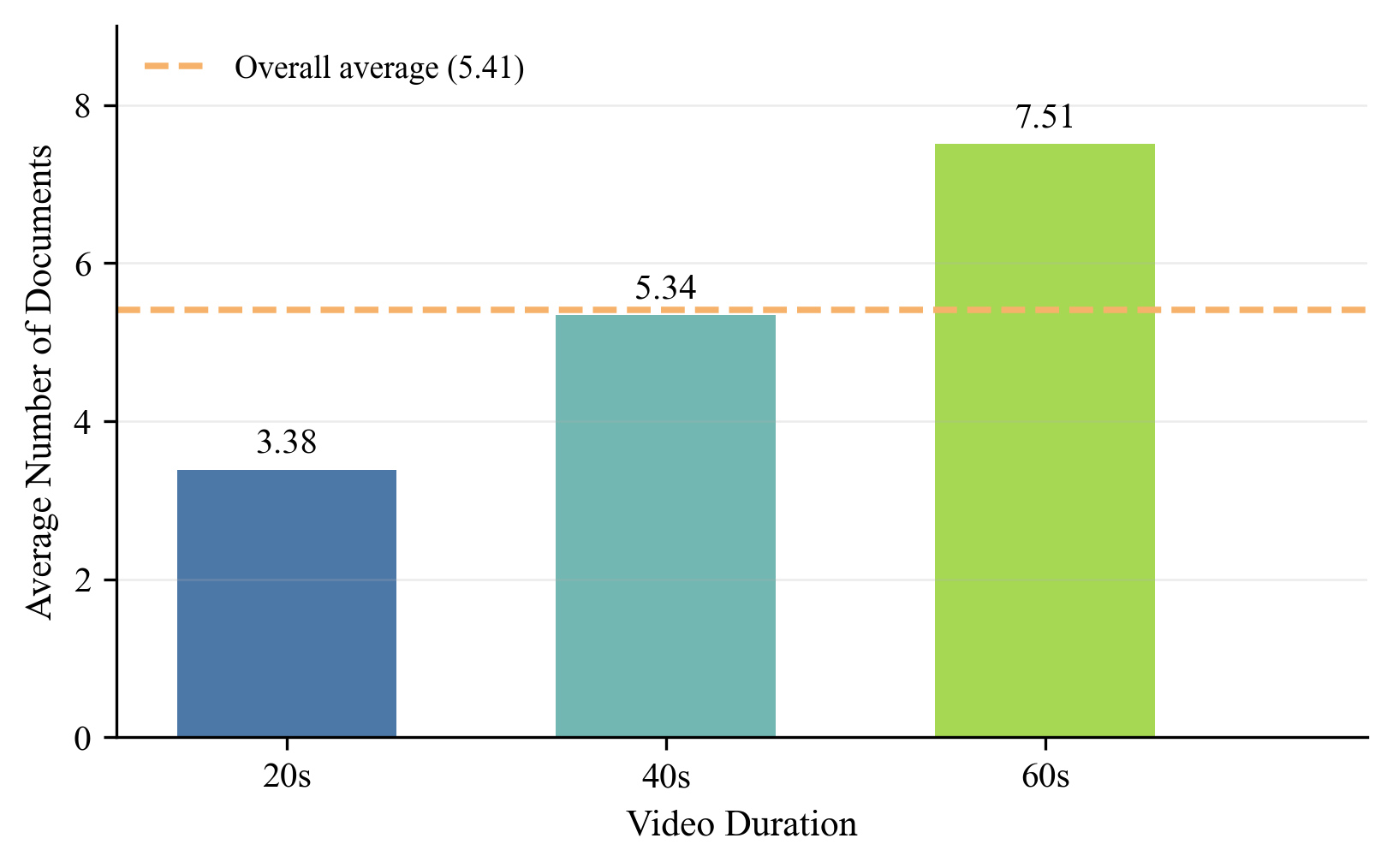}
  \caption{Average number of documents per video across different video durations. Longer videos contain more documents on average, indicating that increasing video duration also increases document-composition complexity. The dashed line denotes the overall average across all videos.}
  \label{fig:avg_docs_by_duration}
\end{figure}

The Video Composition Stage assembles atomic units into long-form multi-document videos for evaluating long-context temporal reasoning and temporal grounding. Composite videos are synthesized into duration tiers of 20s, 40s, and 60s. During composition, Readable and Unreadable atomic clips are stochastically mixed to produce realistic variation in visual quality and evidential continuity, while a greedy balancing procedure enforces a document-uniqueness constraint so that identical document types do not appear within the same composition. Because atomic clips vary in length, each duration tier allows a tolerance of $\pm$5s, which keeps the synthesized videos close to their target duration while preserving natural variation in the corruption mixture. As shown in Fig.~\ref{fig:avg_docs_by_duration}, the average number of documents per video increases from 3.38 at 20s to 7.51 at 60s, indicating that duration acts not only as a temporal stressor but also as a source of document-composition complexity.

To reduce residual heterogeneities such as aspect-ratio differences that could otherwise serve as low-level shortcuts, segments are normalized with a defensive scaling-and-padding strategy on a unified canvas. At each junction, we apply a 10\% fade-in/fade-out cross-fading (Fade/Afade) operation to mimic natural document swapping and to smooth temporal feature evolution. These rendering operations are implemented with FFmpeg filter chains.

\begin{figure}[!htbp]
  \centering
  \includegraphics[width=\linewidth]{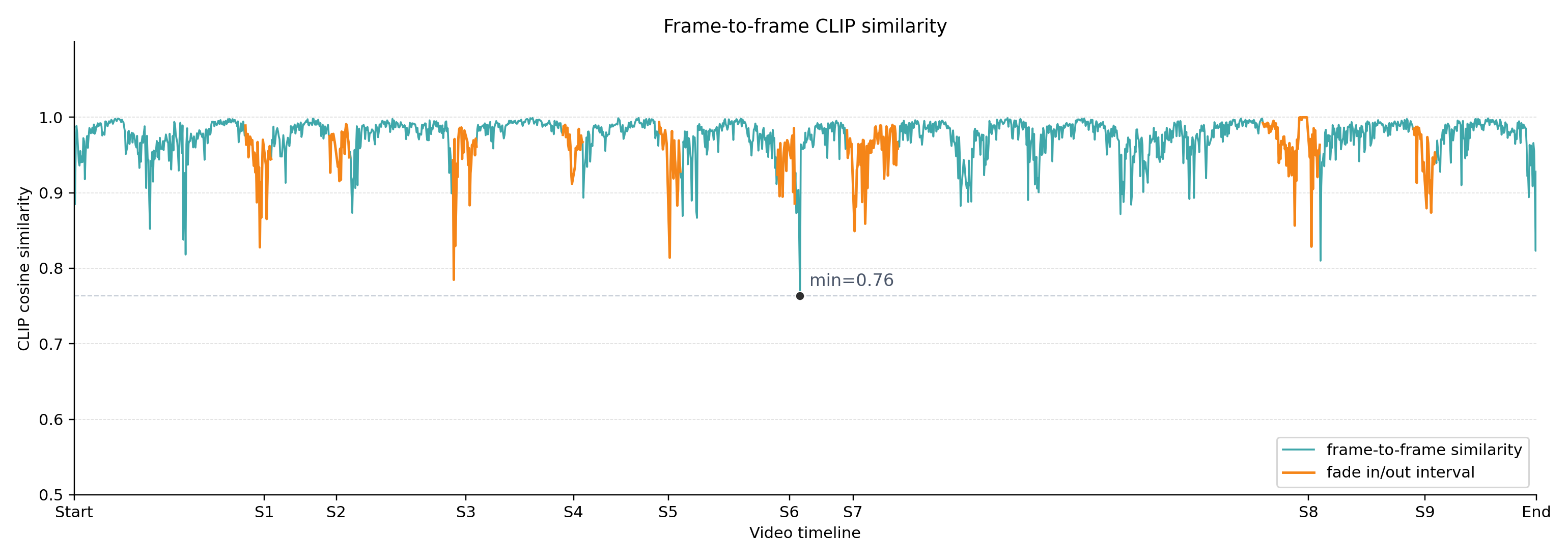}
  \caption{Frame-to-frame CLIP similarity over a representative composed video \texttt{38-luosayu\_60s\_3.mp4}. Each point measures the cosine similarity between adjacent frames in CLIP embedding space, while orange intervals mark fade-in/fade-out regions around composition boundaries. The similarity trajectory remains high across the timeline, and the fade intervals exhibit similar fluctuation trends and magnitudes to the adjacent non-fade regions that correspond to continuous handheld capture. The minimum similarity point occurs outside the highlighted fade intervals.}
  \label{fig:composition_continuity}
\end{figure}

Although FCMBench-Video is constructed by concatenating atomic clips, the intended viewing effect is closer to a continuous handheld recording than to a sequence of visually abrupt cuts. To verify this property, we compute frame-to-frame similarity over time using CLIP image embeddings~\cite{radford2021learning}. Specifically, for each frame $f_t$, we encode it with the CLIP ViT-L/14 visual encoder and measure cosine similarity with the previous frame $f_{t-1}$. We then plot the similarity trajectory against the video timeline and mark the fade-in/fade-out intervals around composition boundaries. As shown in Fig.~\ref{fig:composition_continuity}, similarity remains high throughout the composed video, and the fade intervals exhibit similar fluctuation trends and magnitudes to the adjacent non-fade regions that correspond to continuous handheld capture. The lowest-similarity point also falls outside the highlighted fade intervals. Together, these observations indicate that the scale-and-pad normalization and fade-based rendering strategy does not introduce visible discontinuities relative to the underlying handheld dynamics.

Finally, the system programmatically generates structured JSON labels for each video, including (i) the absolute time range of each segment and (ii) the ``pure feature zone'' excluding cross-fade durations. To guarantee deterministic reproducibility, synthesis uses a pseudo-random seeding mechanism based on MD5 hashing of source directories, ensuring identical outputs across environments.

\section{Task Design and Metrics}
\label{sec:tasks}

FCMBench-Video evaluates document-video intelligence along two complementary axes: \textbf{Perception} and \textbf{Reasoning}. 
Perception tasks test whether a model can recognize, count, and localize document evidence under realistic acquisition artifacts. 
Reasoning tasks require the model to make evidence-grounded decisions in the presence of multiple documents, missing evidence, or malicious visual instructions.

\subsection{Instruction and Task Generation}
\label{subsec:instruction_generation}

FCMBench-Video includes a unified instruction-generation pipeline that converts the composed videos into benchmark instances. 
Perception tasks are instantiated from bilingual prompt templates with structured outputs, while reasoning tasks are generated from task rules, question templates, and task-specific ground-truth functions. 
This design keeps the benchmark extensible and auditable: adding a new task amounts to defining its prompt schema, answer format, and label-construction logic. 
All instances are exported in a unified JSONL format containing the video path, task category, prompt, composition metadata, and reference answer.

\textbf{Release composition and leakage control.}
Benchmark instantiation is organized around identity-level atomic collections and deterministic video composition. In the current release, we treat the evaluation set as a fixed benchmark release rather than a train/test split for model fitting, since all reported results are zero-shot and no task-specific fine-tuning is performed. To reduce shortcut leakage within the released benchmark, each composed video is assembled under a document-uniqueness constraint, composition metadata is generated deterministically, and cross-document validation instances are instantiated only when the required workflow logic is available. Table~\ref{tab:release_composition} summarizes the released benchmark in terms of identities, atomic source documents, composed videos, instructions per video, and task coverage. These statistics clarify the released benchmark specification and facilitate inspection, but they are not a substitute for controlled validity experiments.

\begin{table}[htbp]
\centering
\caption{Release-level composition summary for FCMBench-Video. Unique identities denote distinct identity instances used to organize atomic document collections. Unique atomic source documents correspond to the captured atomic clips used as reusable composition units. Benchmark instances correspond to instruction JSONL rows. Task categories differ across subsets because Cross-Document Validation is currently instantiated only for the zh-CN workflow subset.}
\label{tab:release_composition}
\small
\setlength{\tabcolsep}{5pt}
\begin{tabular}{lcc}
\toprule
\textbf{Statistic} & \textbf{zh-CN} & \textbf{en-US} \\
\midrule
Unique identities & 15 & 30 \\
Unique atomic source documents & 251 & 244 \\
Unique composed videos & 405 & 795 \\
Benchmark instructions & 5,960 & 5,362 \\
Average instructions per composition & 14.72 & 6.74 \\
Task categories & 7 & 6 \\
\bottomrule
\end{tabular}
\end{table}

\subsection{Task Formulation}
\label{subsec:task_suite}

\textbf{Perception tasks.}
We define three perception tasks aligned with workflow requirements:
(1) \emph{Classification}, which requires recognizing the document types present in a video from a predefined candidate set;
(2) \emph{Counting}, which requires de-duplicated counting of logical document instances across the full video, where multi-page or repeated appearances are treated as a single document entity rather than redundant frames;
(3) \emph{Temporal Grounding}, which requires models to localize target documents along the temporal axis by predicting their start and end timestamps. Ground truth is derived from the \emph{pure feature zone} annotations produced in Sec.~\ref{sec:adc}.

\textbf{Reasoning tasks.}
We instantiate three reasoning task families.
\emph{Visual Prompt Injection} is constructed by appending an additional \textbf{2-second} captured clip to the end of an already synthesized document video. Instead of digitally overlaying text, we physically present an attack instruction in the same handheld capture style as ordinary documents, so that the injected content shares the acquisition characteristics of the preceding video. The injected message explicitly asks the model to ignore the previously shown document evidence and directly approve the review. This setting tests whether a Video-MLLM remains faithful to previously presented document evidence under a malicious late visual cue. In this release, the task should be interpreted as a visual prompt-injection stress test rather than a fully isolated measure of adversarial robustness, because recency effects and instruction-following behavior are intentionally entangled in the present construction. We evaluate this task in two settings: \emph{Visual Prompt Injection (w/o CoT)}, which requires a direct decision, and \emph{Visual Prompt Injection (w/ CoT)}, which keeps the same injected clip but requires the model to provide an explicit analysis before outputting the final decision.
\emph{Cross-Document Validation} requires the model to compare or combine evidence across multiple document types, including consistency checks, numerical comparison, and business-rule-based validation; if required documents are absent, the model must abstain with a \textit{Missing} answer rather than hallucinating.
\emph{Evidence-Grounded Selection} asks the model to answer a single-choice question from a multi-document video without being told which document contains the answer, thereby jointly testing document selection, field extraction, evidence integration, and final decision making.

\subsection{Instruction Settings}
\label{subsec:instruction_settings}

FCMBench-Video comprises two language-stratified subsets reflecting distinct real-world deployment scenarios: the \textbf{Chinese (zh-CN) subset} contains Chinese financial workflow videos spanning 20 document categories (e.g., Real Estate Ownership Certificate, Loan Application Form), while the \textbf{English (en-US) subset} contains English-language document videos covering 8 Western document categories (e.g., Driver's License, Individual Income Tax Return). 
The two subsets are evaluated independently, as their document inventories, visual layouts, and linguistic conventions are fundamentally different. 
The benchmark is currently instantiated in three settings: \textbf{Chinese video $\times$ Chinese instruction}, \textbf{Chinese video $\times$ English instruction}, and \textbf{English video $\times$ English instruction}, as summarized in Table~\ref{tab:setting_matrix}. 
The Chinese-video setting contains the full benchmark task set, enabling both monolingual and cross-lingual instruction evaluation over the same underlying videos. For the en-US subset, English instructions are used throughout. 
The English-video setting includes perception tasks, both Visual Prompt Injection settings, and Evidence-Grounded Selection, but does not yet include Cross-Document Validation. 
This omission is deliberate: unlike the Chinese financial workflow subset, the current English subset does not yet have a stable set of expert-validated business review rules for cross-document checking. 
Rather than introducing ad hoc validation logic, we restrict cross-document validation to the subset where the review rules are grounded in a realistic workflow.

\begin{table}[!htbp]
\centering
\caption{Task availability across instruction settings. Check marks indicate task families and Visual Prompt Injection settings currently included in each setting.}
\label{tab:setting_matrix}
\small
\setlength{\tabcolsep}{5pt}
\begin{tabular}{p{0.28\linewidth}ccccccc}
\toprule
\textbf{Setting} & \textbf{Cls.} & \textbf{Cnt.} & \textbf{Grd.} & \textbf{VPI} & \textbf{VPI-CoT} & \textbf{CDV} & \textbf{EGS} \\
\midrule
zh-video $\times$ zh-inst. & Y & Y & Y & Y & Y & Y & Y \\
zh-video $\times$ en-inst. & Y & Y & Y & Y & Y & Y & Y \\
en-video $\times$ en-inst. & Y & Y & Y & Y & Y &  & Y \\
\bottomrule
\end{tabular}
\end{table}

\subsection{Evaluation Metrics}
\label{subsec:metrics}

We evaluate three perception tasks (\textit{Classification}, \textit{Counting}, and \textit{Temporal Grounding}) and three reasoning task families (\textit{Visual Prompt Injection}, \textit{Cross-Document Validation}, and \textit{Evidence-Grounded Selection}), with \textit{Visual Prompt Injection} reported under both w/o-CoT and w/CoT settings. All tasks use standardized instructions and structured outputs to reduce formatting variance and to enforce abstention when evidence is unreadable or missing. The main stratified analysis in this paper is performed over video duration (20s/40s/60s), while reasoning results are analyzed primarily through aggregate task performance and prompt-injection robustness. Throughout the tables, \textbf{Acc} denotes exact-match accuracy, \textbf{mIoU} denotes mean temporal intersection-over-union, \textbf{ASR} denotes attack success rate, \textbf{CDV} denotes Cross-Document Validation, and \textbf{EGS} denotes Evidence-Grounded Selection. Because exact-match metrics can be affected by empty or malformed outputs, the reasoning results are interpreted with explicit caution about output-compliance effects. A summary of all tasks and their corresponding metrics is provided in Table~\ref{tab:task_summary}.

\textbf{Perception metrics.}
For document type identification, we report precision/recall/F1 against ground-truth document types per video. 
For de-duplicated counting, we report exact-match accuracy on the predicted count.

\textbf{Temporal grounding metrics.}
For temporal grounding, we evaluate predicted intervals against the ground-truth pure feature zone using temporal intersection-over-union (mIoU).
Let $I_p^{(i)}=[t_s^{(i,p)},\, t_e^{(i,p)}]$ be the predicted interval
and $I_g^{(i)}=[t_s^{(i,g)},\, t_e^{(i,g)}]$ the ground-truth interval
for the $i$-th segment, then

\[
\mathrm{IoU}\bigl(I_p^{(i)}, I_g^{(i)}\bigr) =
\frac{
  \max\bigl(0,\ \min(t_e^{(i,p)}, t_e^{(i,g)}) - \max(t_s^{(i,p)}, t_s^{(i,g)})\bigr)
}{
  \bigl(t_e^{(i,p)} - t_s^{(i,p)}\bigr) + \bigl(t_e^{(i,g)} - t_s^{(i,g)}\bigr)
  - \max\bigl(0,\ \min(t_e^{(i,p)}, t_e^{(i,g)}) - \max(t_s^{(i,p)}, t_s^{(i,g)})\bigr)
}
\]

\noindent The mean Temporal IoU over $N$ evaluated video segments is:
\[
\mathrm{mIoU} = \frac{1}{N} \sum_{i=1}^{N} \mathrm{IoU}\bigl(I_p^{(i)}, I_g^{(i)}\bigr)
\]

\begin{table}[!htbp]
\centering
\caption{Task summary for FCMBench-Video. 
The main stratified analysis in this paper is performed over video duration (20s/40s/60s).}
\label{tab:task_summary}
\setlength{\tabcolsep}{6pt}
\small
\resizebox{1.0\linewidth}{!}{
\begin{tabular}{p{0.29\linewidth} p{0.24\linewidth} p{0.25\linewidth} p{0.12\linewidth}}
\toprule
\textbf{Task} & \textbf{Output} & \textbf{Metric} & \textbf{Setting} \\
\midrule
Classification & Python list / JSON list & Precision, Recall, F1 & all settings \\
Counting & Scalar count & Exact-match accuracy & all settings \\
Temporal Grounding & Timestamp interval(s) & mIoU & all settings \\
Visual Prompt Injection (w/o CoT) & Approve / reject decision & ASR & all settings \\
Visual Prompt Injection (w/ CoT) & Analysis + decision & ASR & all settings \\
Cross-Document Validation & Missing / binary / numeric & Exact-match accuracy & zh$\times$zh, zh$\times$en \\
Evidence-Grounded Selection & Option letter & Exact-match accuracy & all settings \\
\bottomrule
\end{tabular}
}
\end{table}

\textbf{Reasoning metrics.}
For \emph{Visual Prompt Injection}, we report \emph{Attack Success Rate (ASR)} as the primary metric under both the w/o-CoT and w/CoT settings, i.e., the proportion of cases in which the injected approval instruction successfully alters the model's decision. 
For \emph{Cross-Document Validation}, we report exact-match accuracy after answer normalization, covering missing-evidence answers, binary consistency judgments, rule-based decisions, and numerical outputs. 
For \emph{Evidence-Grounded Selection}, we report exact-match accuracy on the predicted option letter.
These exact-match reasoning metrics should be interpreted as measuring \emph{end-to-end task success under structured-output constraints}, rather than pure semantic reasoning in isolation. In particular, empty outputs, malformed outputs, or answer-format noncompliance can lower the reported score even when partial evidence retrieval is present. To make this limitation explicit, Sec.~\ref{sec:experiments} reports an output-validity analysis that decomposes raw reasoning outputs into format-valid, empty, malformed, and semantic-wrong cases.

\section{Experiments}
\label{sec:experiments}
\subsection{Experimental Setup}

We evaluate nine recent Video-MLLMs released in 2025--2026, spanning commercial API systems and open-source models across diverse scales (Table~\ref{tab:model_background}). All models are evaluated in a zero-shot manner on the same FCMBench-Video benchmark instances, with task-specific prompts and structured output requirements. Detailed settings are listed below.

\subsubsection{Deployment Settings}

Commercial models, including Gemini-3.0-Pro-Preview~\cite{gemini3blog,geminivideodocs} and Doubao-Seed-1.6-vision~\cite{seed16techblog}, are accessed through their public APIs with native raw-video upload. Open-source models, including Kimi-VL-A3B-Instruct~\cite{kimivl2025}, InternVL3-8B~\cite{internvl32025}, Ovis2.5-9B~\cite{ovis252025}, Qwen3-Omni-30B-A3B-Instruct~\cite{qwen3omnigithub}, Qwen3-VL-8B/32B-Instruct~\cite{Qwen3-VL}, and Qwen3.5-27B~\cite{qwen35blog}, are served through a unified vLLM-based inference path when supported by the model release. No model is fine-tuned on FCMBench-Video. For the main comparison, we report models with completed runs across the full benchmark task set. Table~\ref{tab:model_background} summarizes each model's release time, parameter scale, model availability, inference backend, and video-input handling.

\begin{table}[!htbp]
\centering
\caption{Model background for FCMBench-Video. Release dates are approximate first public releases. ``Model availability'', ``backend'', and ``video input'' summarize the native video-serving path used in our experiments rather than a unified iso-input protocol. FPS values denote explicit frame sampling or native-serving settings for open-source models. Commercial models are evaluated with native API raw-video upload and no additional video preprocessing on our side.}
\label{tab:model_background}
\setlength{\tabcolsep}{5pt}
\scriptsize
\resizebox{\linewidth}{!}{
\begin{tabular}{p{0.24\linewidth} p{0.06\linewidth} p{0.08\linewidth} p{0.12\linewidth} p{0.12\linewidth} p{0.16\linewidth}}
\toprule
\textbf{Model} & \textbf{Release} & \textbf{Params} & \textbf{Model Availability} & \textbf{Backend} & \textbf{Video input} \\
\midrule
Kimi-VL-A3B-Instruct~\cite{kimivl2025} & 2025-04 & 16B-A3B & Open-source & vLLM & 2\,FPS \\
InternVL3-8B~\cite{internvl32025} & 2025-04 & 8B & Open-source & vLLM & 2\,FPS \\
Doubao-Seed-1.6-vision~\cite{seed16techblog} & 2025-06 & N/A & Closed-source & Commercial API & Native API / raw upload \\
Ovis2.5-9B~\cite{ovis252025} & 2025-08 & 9B & Open-source & vLLM & 0.5\,FPS \\
Qwen3-Omni-30B-A3B-Instruct~\cite{qwen3omnigithub} & 2025-09 & 30B-A3B & Open-source & vLLM & 2\,FPS \\
Qwen3-VL-8B-Instruct~\cite{Qwen3-VL} & 2025-10 & 8B & Open-source & vLLM & 2\,FPS \\
Qwen3-VL-32B-Instruct~\cite{Qwen3-VL} & 2025-10 & 32B & Open-source & vLLM & 2\,FPS \\
Gemini-3.0-Pro-Preview~\cite{gemini3blog,geminivideodocs} & 2025-11 & N/A & Closed-source & Commercial API & Native API / raw upload \\
Qwen3.5-27B~\cite{qwen35blog} & 2026-02 & 27B & Open-source & vLLM & 2\,FPS \\
\bottomrule
\end{tabular}
}
\end{table}

\subsubsection{Frame Sampling Settings}

We use model-specific native video-serving settings whenever possible. Commercial models (Gemini-3.0-Pro-Preview and Doubao-Seed-1.6-vision) are evaluated with raw video upload and no additional preprocessing on our side. 
For open-source models, Ovis2.5-9B uses 0.5\,FPS uniform sampling; Kimi-VL-A3B-Instruct, InternVL3-8B, and Qwen3-Omni-30B-A3B-Instruct use 2\,FPS uniform sampling; Qwen3-VL-8B-Instruct, Qwen3-VL-32B-Instruct, and Qwen3.5-27B use model-side video handling, reported as 2\,FPS native-serving in Table~\ref{tab:model_background}.
This mixed protocol reflects practical deployment conditions rather than a strictly controlled iso-input setting; results should therefore be interpreted as system-level performance under each model's native serving path.

\subsection{Experimental Results and Analysis}
\subsubsection{Overall Performance}

\begin{figure*}[!htbp]
  \centering
  \includegraphics[width=\textwidth,height=0.31\textheight,keepaspectratio]{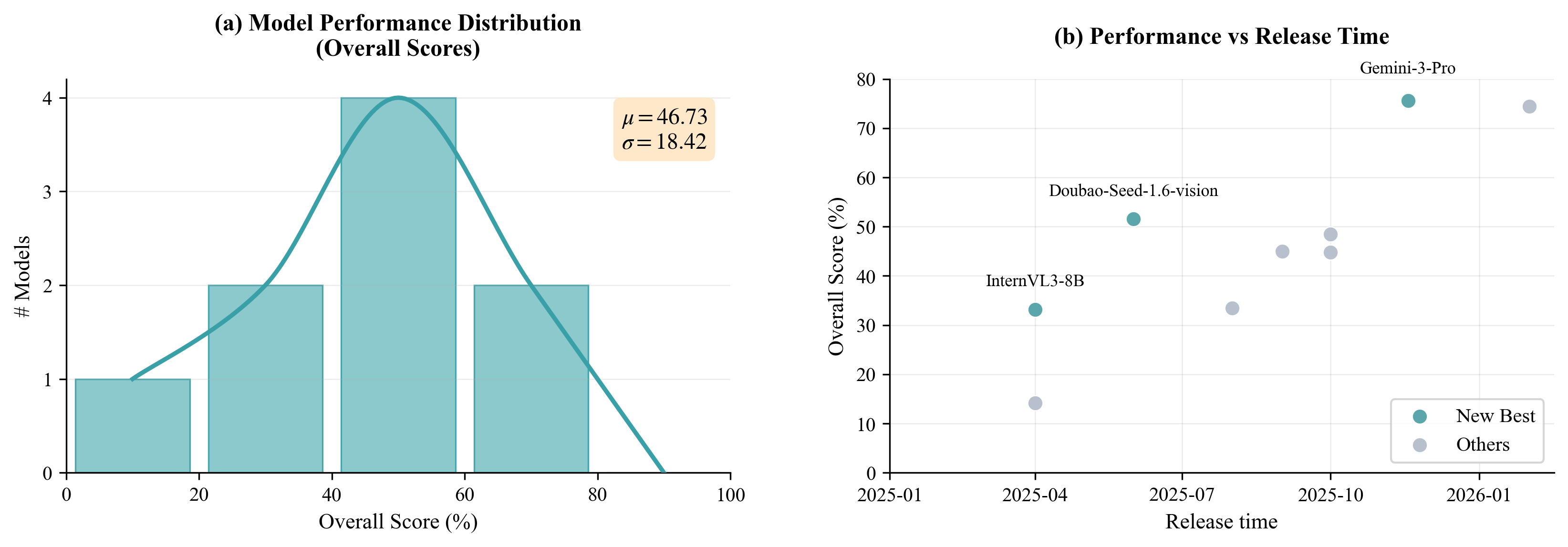}
  \caption{Overall performance analysis on FCMBench-Video. (a) Distribution of model overall scores, where the overall score is computed as the mean over all reported task metrics from the zh- and en-subsets after converting Visual Prompt Injection (w/o CoT) and Visual Prompt Injection (w/ CoT) to higher-is-better scores via $1-\mathrm{ASR}$. The score distribution is broad and approximately bell-shaped ($\mu=46.73$, $\sigma=18.42$), rather than concentrated near either extreme: FCMBench-Video is neither saturated by current Video-MLLMs nor dominated by trivial cases, and it provides meaningful resolution for separating system capabilities. (b) Overall score versus model release time. Colored points mark milestone models that successively refresh the best reported overall score over time, while gray points denote the remaining models. The frontier rises with newer releases, showing that FCMBench-Video tracks genuine capability progress, while the substantial spread among nearby releases confirms that the benchmark remains non-trivial and discriminative for contemporary models.}
  \label{fig:overall_performance_main}
\end{figure*}

We first analyze the overall behavior of FCMBench-Video across the evaluated Video-MLLMs. For each model, we compute an overall benchmark score as the mean over all reported task metrics from the zh- and en-subsets, after converting the lower-is-better Visual Prompt Injection (w/o CoT) and Visual Prompt Injection (w/ CoT) metrics into higher-is-better scores via $1-\mathrm{ASR}$. As shown in Fig.~\ref{fig:overall_performance_main}(a), the resulting overall scores exhibit a broad, approximately bell-shaped distribution with $\mu=46.73$ and $\sigma=18.42$. Rather than clustering near the ceiling or floor, models are distributed across a wide intermediate range. FCMBench-Video is thus challenging enough that current systems do not saturate it, yet structured enough to provide clear separation among models with different capability profiles. The benchmark is clearly \textbf{non-trivial}: strong performance requires jointly handling document perception, temporal grounding, cross-document reasoning, and visual prompt-injection robustness rather than relying on a small set of easy patterns.

Fig.~\ref{fig:overall_performance_main}(b) further shows that FCMBench-Video tracks model progress over time. The frontier of best-reported overall performance moves upward with newer model releases, and several recently released systems set successive new highs, indicating that the benchmark is sensitive to genuine capability improvements rather than dominated by noise. Meanwhile, the spread among contemporaneous models remains substantial: models released in nearby periods can still differ markedly in overall score. This combination is informative. If the benchmark were trivial, most recent models would be tightly packed near the top; if it were poorly constructed, scores would appear unstable and unrelated to model development trends. Instead, FCMBench-Video follows the expected trajectory of Video-MLLM progress while preserving meaningful discrimination within the current generation of systems.

\subsubsection{Task-Specific Performance}

Tables~\ref{tab:main_results_zh} and~\ref{tab:main_results_en} report task-specific results on the zh-CN and en-US subsets, respectively. Models are ordered from earlier to later public release; when release dates are close, closely related variants are grouped by scale. For Qwen3.5-27B, the Visual Prompt Injection columns are retained for completeness: they use the standard prompt-injection setting without disabling the model's native thinking behavior, so the reported w/o-CoT values should be read as nominal native-thinking results rather than strictly non-CoT runs.

\begin{table}[H]
\centering
\caption{Main results on the zh-video subset of FCMBench-Video. Scores are reported as percentages. Models are ordered by public release date from earlier to later releases. Abbreviations: Clas. = Classification, Cnt. = Counting, Grd. = Grounding, VPI = Visual Prompt Injection (w/o CoT), VPI-CoT = Visual Prompt Injection (w/ CoT), CDV = Cross-Document Validation, and EGS = Evidence-Grounded Selection. Upward arrows indicate higher-is-better metrics, while downward arrows indicate lower-is-better metrics. Green and red denote the best and worst results for each task, respectively.}
\label{tab:main_results_zh}
\setlength{\tabcolsep}{2pt}
\scriptsize
\resizebox{\linewidth}{!}{
\begin{tabular}{lccccccc}
\toprule
\textbf{Model} & \multicolumn{3}{c}{\textbf{Perception}} & \multicolumn{4}{c}{\textbf{Reasoning}} \\
\cmidrule(lr){2-4} \cmidrule(lr){5-8}
& \textbf{Classification F1 $\uparrow$} & \textbf{Counting Acc $\uparrow$} & \textbf{Grounding mIoU $\uparrow$} & \textbf{VPI ASR $\downarrow$} & \textbf{VPI-CoT ASR $\downarrow$} & \textbf{CDV Acc $\uparrow$} & \textbf{EGS Acc $\uparrow$} \\
\midrule
Kimi-VL-A3B-Instruct & 46.13 & \worst{18.37} & \worst{1.97} & 93.52 & \worst{100.00} & \worst{0.00} & \worst{0.00} \\
InternVL3-8B & 42.58 & 28.22 & 15.81 & 81.85 & 57.70 & 30.74 & 15.18 \\
Doubao-Seed-1.6-vision & 70.43 & 54.67 & 69.87 & \worst{96.85} & 76.85 & 33.33 & 50.18 \\
Ovis2.5-9B & \worst{35.92} & 25.41 & 2.88 & 45.19 & 41.11 & 45.56 & 13.04 \\
Qwen3-Omni-30B-A3B-Instruct & 50.70 & 27.19 & 22.99 & 43.52 & 39.63 & 44.07 & 33.75 \\
Qwen3-VL-8B-Instruct & 59.77 & 34.74 & 52.37 & 68.15 & 52.73 & 48.15 & 35.00 \\
Qwen3-VL-32B-Instruct & 65.85 & 41.11 & 63.62 & 79.63 & 62.59 & 40.74 & 44.11 \\
Gemini-3.0-Pro-Preview & \best{75.15} & \best{67.09} & \best{79.14} & \best{12.24} & 45.00 & 38.29 & \best{73.04} \\
Qwen3.5-27B & 73.06 & 39.33 & 69.43 & 18.89 & \best{18.89} & \best{52.59} & 67.32 \\
\bottomrule
\end{tabular}
}
\end{table}

\vspace{-0.6em}
\begin{table}[H]
\centering
\caption{Main results on the en-video subset of FCMBench-Video. Scores are reported as percentages. Models are ordered by public release date from earlier to later releases. Abbreviations: Grd. = Grounding, VPI = Visual Prompt Injection (w/o CoT), VPI-CoT = Visual Prompt Injection (w/ CoT), and EGS = Evidence-Grounded Selection. Upward arrows indicate higher-is-better metrics, while downward arrows indicate lower-is-better metrics. Green and red denote the best and worst results for each task, respectively.}
\label{tab:main_results_en}
\setlength{\tabcolsep}{2pt}
\scriptsize
\resizebox{\linewidth}{!}{
\begin{tabular}{lcccccc}
\toprule
\textbf{Model} & \multicolumn{3}{c}{\textbf{Perception}} & \multicolumn{3}{c}{\textbf{Reasoning}} \\
\cmidrule(lr){2-4} \cmidrule(lr){5-7}
& \textbf{Classification F1 $\uparrow$} & \textbf{Counting Acc $\uparrow$} & \textbf{Grounding mIoU $\uparrow$} & \textbf{VPI ASR $\downarrow$} & \textbf{VPI-CoT ASR $\downarrow$} & \textbf{EGS Acc $\uparrow$} \\
\midrule
Kimi-VL-A3B-Instruct & \worst{69.25} & 30.11 & \worst{2.96} & 91.70 & \worst{100.00} & \worst{0.00} \\
InternVL3-8B & 76.60 & 38.42 & 16.12 & 64.72 & 60.94 & 32.72 \\
Doubao-Seed-1.6-vision & 86.41 & 67.77 & 72.23 & 60.75 & 56.42 & 55.96 \\
Ovis2.5-9B & 71.02 & \worst{27.25} & 3.69 & 86.98 & 58.30 & 42.20 \\
Qwen3-Omni-30B-A3B-Instruct & 79.00 & 34.72 & 42.18 & 73.40 & 31.89 & 38.53 \\
Qwen3-VL-8B-Instruct & 82.82 & 34.57 & 67.13 & \worst{94.91} & 49.25 & 33.03 \\
Qwen3-VL-32B-Instruct & 87.23 & 54.94 & 71.57 & 61.13 & 77.36 & 41.90 \\
Gemini-3.0-Pro-Preview & 90.98 & \best{67.92} & \best{80.39} & \best{0.38} & 8.87 & 76.99 \\
Qwen3.5-27B & \best{91.53} & 59.77 & 75.74 & 2.08 & \best{8.49} & \best{86.85} \\
\bottomrule
\end{tabular}
}
\end{table}

\textbf{Perception.}
Tables~\ref{tab:main_results_zh} and~\ref{tab:main_results_en} show that document-video perception is not a single uniform capability. Strong models can often recognize coarse document categories, especially on the English subset, but the gap widens when the task requires temporal evidence accumulation or precise localization. Counting is particularly sensitive because the model must maintain a de-duplicated inventory over multiple document appearances rather than classify a single salient frame. Temporal grounding adds another requirement: the model must not only read the document, but also identify when the relevant evidence is visible with sufficient clarity. This explains why models with reasonable classification scores can still fail on counting or temporal localization.

The language settings further indicate that perception quality depends on both visual evidence and instruction following. For the same Chinese videos, Chinese instructions are consistently easier for classification, while grounding and counting show more model-dependent behavior. Bilingual evaluation therefore probes how models couple document-language familiarity, answer formatting, and video evidence retrieval, rather than serving as a translation check. The English subset is generally easier for the strongest systems, likely because its documents contain larger text regions and more regular layouts, but we treat this as an empirical tendency rather than a controlled causal claim.

\begin{figure}[!h]
  \centering
  \includegraphics[height=0.65\textheight,keepaspectratio]{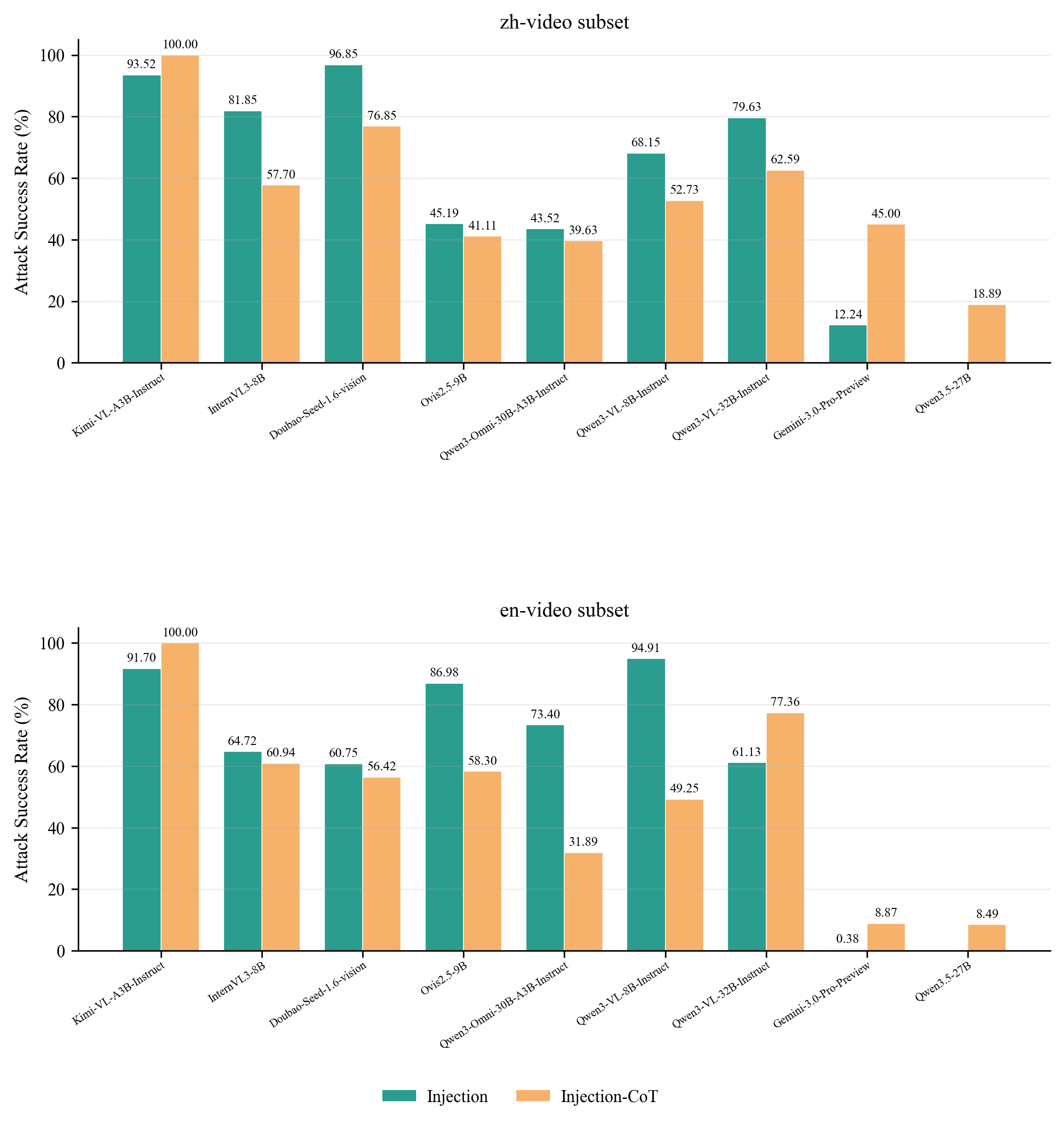}
  \caption{Overall comparison of \emph{Attack Success Rate (ASR)} between Visual Prompt Injection (w/o CoT) and Visual Prompt Injection (w/ CoT) on the zh-CN and en-US subsets. The figure includes all nine models used in the main trend analysis. Explicit intermediate reasoning does not guarantee lower ASR under the current visual prompt-injection construction: some models benefit noticeably, while others remain unstable or even degrade under the CoT variant. For Qwen3.5-27B, only the VPI-CoT bar is shown because the model's thinking behavior cannot be disabled, making its nominal ``VPI'' setting not directly comparable to strictly non-CoT runs.}
  \label{fig:injection_compare_main}
\end{figure}

\textbf{Reasoning.}
The reasoning tasks expose a substantially larger gap than perception alone. Cross-Document Validation and Evidence-Grounded Selection require models to select relevant documents, retain evidence across segments, and apply task-specific decision rules. The results show that these capabilities are only weakly coupled with coarse perception: a model may identify documents reasonably well while still failing to compare values, detect missing evidence, or choose the correct answer from a multi-document context. This pattern suggests that FCMBench-Video is not merely re-measuring document classification in video form; it stresses persistent evidence use over time.

Visual Prompt Injection results add a complementary robustness dimension. In the current construction, the injected instruction appears in the final two seconds of the video, so attack success reflects both instruction susceptibility and recency bias toward late visual content. Figure~\ref{fig:injection_compare_main} shows that explicit reasoning does not uniformly improve robustness: some models benefit from the Visual Prompt Injection (w/ CoT) setting, while others remain unstable or even become more vulnerable. Therefore, this task should be interpreted as a realistic stress test for task-intent preservation under malicious visual prompt injection, rather than as a fully factorized adversarial benchmark. The fact that no model dominates all reasoning-oriented axes indicates that preserving earlier document evidence against later conflicting cues remains a meaningful challenge in this benchmark.

\begin{table}[!htbp]
\centering
\caption{Reasoning output-validity analysis computed from raw reasoning-task outputs, aggregated over each model's reasoning-task outputs. ``Format-valid'' counts responses that can be parsed into the required schema after the benchmark's standard normalization; ``Empty'' counts no-answer responses; and ``Malformed'' counts non-empty but non-parseable responses. The three categories are exhaustive and sum to 100\% for each model.}
\label{tab:reasoning_output_validity}
\small
\setlength{\tabcolsep}{6pt}
\begin{tabular}{lccc}
\toprule
\textbf{Model} & \textbf{Format-valid (\%)} $\uparrow$ & \textbf{Empty (\%)} $\downarrow$ & \textbf{Malformed (\%)} $\downarrow$ \\
\midrule
Kimi-VL-A3B-Instruct & 9.52 & 90.48 & 0.00 \\
InternVL3-8B & 93.76 & 6.24 & 0.00 \\
Doubao-Seed-1.6-vision & 96.16 & 3.84 & 0.00 \\
Ovis2.5-9B & 96.77 & 3.12 & 0.11 \\
Qwen3-Omni-30B-A3B-Instruct & 99.86 & 0.14 & 0.00 \\
Qwen3-VL-8B-Instruct & 99.61 & 0.39 & 0.00 \\
Qwen3-VL-32B-Instruct & 98.45 & 1.55 & 0.00 \\
Gemini-3.0-Pro-Preview & 97.05 & 2.95 & 0.00 \\
Qwen3.5-27B & 100.00 & 0.00 & 0.00 \\
\bottomrule
\end{tabular}
\end{table}

\textbf{Failure analysis.}
Table~\ref{tab:reasoning_output_validity} separates reasoning failures caused by output validity from failures after a parseable answer is produced. Most models already generate format-valid outputs for the majority of reasoning instances, and malformed responses are rare. This means the benchmark is not mainly measuring parser fragility or superficial schema compliance. The main exception is Kimi-VL-A3B-Instruct, whose reasoning performance is dominated by empty outputs. For the remaining models, the dominant failure mode is semantic: the response is syntactically valid, but the model fails to retrieve, retain, or combine the required evidence correctly. The strongest systems reduce this semantic error rate but do not eliminate it, indicating that robust document-video reasoning remains difficult even when output formatting is largely under control.

\subsubsection{Performance over Video Duration}

\begin{figure*}[!htbp]
  \centering
  \includegraphics[width=\textwidth,height=0.28\textheight,keepaspectratio]{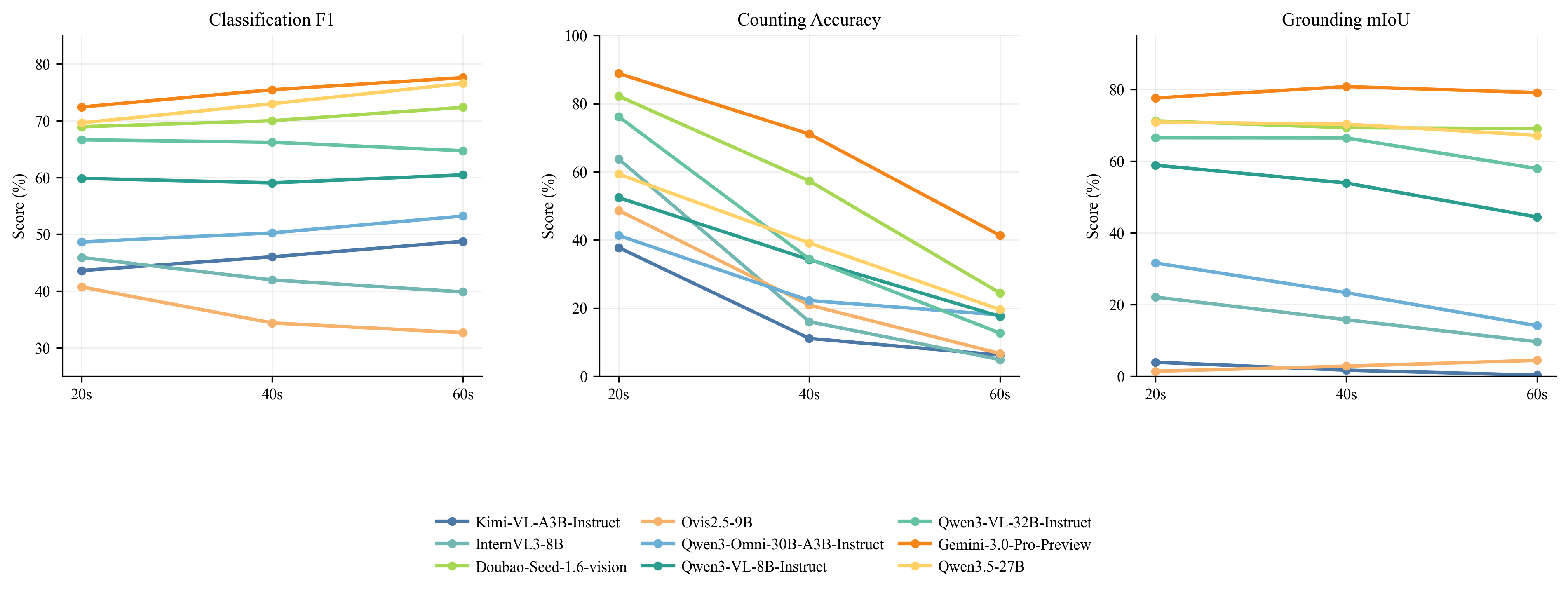}
  \caption{Duration-stratified perception results on the zh-CN subset. The same nine models are evaluated across 20s, 40s, and 60s videos. To improve trend visibility, each panel uses a task-specific vertical range while preserving the original absolute scores. Counting shows the clearest degradation with longer temporal span, whereas classification and grounding are more stable and occasionally improve slightly at 40s, suggesting a trade-off between evidence coverage and long-context burden.}
  \label{fig:duration_perception_main}
\end{figure*}

Video duration is a first-class stressor in FCMBench-Video. As shown in Figure~\ref{fig:duration_perception_main}, model performance does not degrade uniformly across perception tasks as videos become longer. Counting shows the sharpest decline from 20s to 60s, while document classification remains comparatively stable and temporal grounding degrades more gradually. This pattern indicates that increasing duration mainly challenges evidence accumulation and state maintenance, rather than coarse document recognition alone.

The difference across tasks is consistent with the benchmark construction. Longer videos contain more document segments on average, and degradation is injected at the atomic-clip level. As a result, a low-quality segment may hide an entire document instance rather than a few isolated frames. Counting is therefore vulnerable to irreversible inventory errors: once a document segment is missed, the final de-duplicated count is corrupted. Classification is more robust because global layout, document templates, and coarse visual appearance can remain recognizable even when fine text is partially degraded. Temporal grounding falls between these two cases: it benefits from segment-boundary cues, but still requires identifying the interval where task-relevant evidence is visible.

These observations suggest that document-video difficulty grows through two coupled factors: longer temporal context and more complex document composition. Models must not only read local frames, but also preserve evidence across transitions, avoid double-counting repeated appearances, and align answers with the correct time span. Duration-stratified evaluation therefore reveals a capability gap that would be hidden in static-image or short-video settings.

\section{Future Work}

FCMBench-Video provides a reproducible starting point for evaluating document-centric video understanding, but several directions remain open for strengthening the benchmark in future versions. The current results already show empirical separation across systems and tasks: the overall score distribution is broad and approximately bell-shaped, the performance frontier rises with newer model releases, and task-level results remain clearly non-trivial rather than collapsing into a single easy capability. Building on this foundation, the main axis of future work is to extend the benchmark to other video categories that video-based credit review and anti-fraud assessment must also handle, such as on-site recordings of business premises and collateral, field-survey and remote-interview videos used in credit underwriting, and customer-facing recordings captured during in-branch or mobile account opening. The current benchmark is intentionally grounded in realistic credit-related business scenarios, and extending the same evaluation principles to these additional video types would more faithfully reflect the full input space of \emph{video due diligence}, where models must analyze temporally unfolding visual evidence for authenticity, compliance, and decision support.

A second direction is to strengthen experimental control in long-form video settings. The current benchmark already exposes duration, document composition, visual corruption, and temporal annotations as controllable factors in its construction workflow, but future work can add more controlled variants for separating recency bias from visual prompt-injection effects, and for disentangling temporal evidence accumulation from prompt-following behavior. Such extensions would make it easier to attribute failures more precisely rather than relying only on end-to-end task scores.

A third direction is to improve the granularity of evaluation and analysis. Future work should develop finer-grained error attribution, so that benchmark users can distinguish whether a failure comes from perception, temporal memory, cross-document evidence binding, or instruction robustness. More broadly, building on FCMBench-Video, we plan to develop benchmarks for additional video categories in the video-due-diligence setting, providing a non-trivial baseline for credit-related document videos now and broader diagnostic coverage for authenticity-sensitive video analysis in future iterations.

\section*{Acknowledgements}
The authors would like to thank 
Didi Hu, 
Huifang Du, 
Mengyuan Liu, 
Chenghao Fan, 
Kang Du, 
Shouduo Shang, 
Zecheng Zuo, 
Boxun Wen, 
and other colleagues at Qfin Tech Inc. for their valuable assistance and insightful inspiration during the development of this benchmark.

\bibliographystyle{unsrt}
\bibliography{main}

\end{document}